\documentclass[10pt,letterpaper]{article}

\usepackage{cogsci}

\cogscifinalcopy

\usepackage{pslatex}
\usepackage{apacite}
\usepackage{float} 

\usepackage{amsmath}
\usepackage{amsfonts}

\usepackage{graphicx}
\usepackage{caption}
\usepackage{subcaption}

\usepackage{verbatim}

\title{Modeling Communication to Coordinate Perspectives in Cooperation}

\author{
    \begin{tabular}{c c c c}
        \bf Stephanie Stacy$^{1}$ \quad\quad & \bf Chenfei Li$^{1}$  \quad\quad & \bf Minglu Zhao$^{1}$ \quad\quad & \bf Yiling Yun$^{1}$ \\
        \normalfont stephaniestacy@g.ucla.edu \quad\quad & \normalfont adslchf626@g.ucla.edu \quad\quad & \normalfont minglu.zhao@g.ucla.edu \quad\quad & \normalfont yilingsophie@g.ucla.edu
    \end{tabular}
    \\\vspace{-9pt}\\
    \begin{tabular}{c c c}
        \bf Qingyi Zhao$^{3}$ \quad\quad & \bf Max Kleiman-Weiner$^{4,5}$ \quad\quad & \bf Tao Gao$^{1,2}$ \\
        \normalfont zhaoqy1997@g.ucla.edu & \normalfont maxhkw@gmail.com & \normalfont tao.gao@stat.ucla.edu
    \end{tabular}
    \\\vspace{-9pt}\\
    \begin{tabular}{c c c}
        $^1$ Department of Statistics, UCLA & $^2$ Department of Communication, UCLA & $^3$ Department of Computer Science, UCLA 
    \end{tabular}
    \\\vspace{-11pt}\\
    \begin{tabular}{c c}
        $^4$ Department of Brain and Cognitive Sciences, MIT & $^5$ Department of Psychology, Harvard University
    \end{tabular}
}

\begin{document}
\maketitle

\begin{abstract}
Communication is highly overloaded. Despite this, even young children are good at leveraging context to understand ambiguous signals. We propose a computational account of overloaded signaling from a shared agency perspective which we call the Imagined We for Communication. Under this framework, communication is a way for cooperators to coordinate their perspectives, allowing them to act together to achieve shared goals. We assume agents are rational, utility maximizing cooperators, which puts constraints on how signals can be sent and interpreted. We implement this model in a set of simulations which demonstrate this model’s success under increasing ambiguity as well as increasing layers of reasoning. Our model is capable of improving performance with deeper recursive reasoning; however, it outperforms comparison baselines at even the shallowest level of reasoning, highlighting how shared knowledge and cooperative logic can do much of the heavy-lifting in language.

\end{abstract}

\section{Introduction}
Human communication is highly overloaded, conveying rich meaning through sparse, ambiguous signals. For example: while two people are sitting at a table, one person exclaims ``the glass!’’ the other instantly moves her glass away from a precarious spot at the table's edge. Here, ``glass’’ is sparse, leaving the listener to reason why the glass is relevant and how to respond. Additionally, ``glass’’ is ambiguous, it might refer to the eyeglasses safely resting on the table in one context, or the request for a refill in the next. These simple, everyday exchanges involve spontaneity and extreme indirectness, highlighting that humans are intelligent ``inference-making machines’’ \cite{sacks1985inference}. To understand the meaning of a signal in its full context, humans rely on their knowledge of the world, observability of the visual environment, and the (potential) behavior of their partner.

Our work is built on the rich tradition of treating language acquisition as social cognition \cite{bruner1975ontogenesis, grosse2010, tomasello2010origins}. 
Leveraging these empirical insights as well as current computational work, we propose a model capable of cooperative, human-like communication with three novel properties. (1) Joint reasoning using a shared ``We’’ perspective for communication. This allows cooperators to understand signals in terms of what is relevant to everyone (I’m referring to the glass's physical instability, not its shape or content). (2) Joint planning in the physical and visual environment as a constraint on communication. Signals are constrained to interpretations that are expected to improve We’s utility under the cooperative logic that everyone is treated as an equal (``glass'' is not the one closer to you -- that would be your responsibility to save). (3) Indirectness of speech, for rich, multi-dimensional inferences. Signals may tell you \textit{what}, but can also give information about \textit{why} or \textit{how}. (``glass’’ is not referring to the identity of the glass, but an action -- save it).

\section{Cooperation Under an Imagined We}
We advocate for a shared agency model of cooperation: collaborators must be committed to achieving a joint intention as a collective body \cite{gilbert2013joint}. To model this, it is useful to begin by modeling ideal cooperation; here, a single, central ``We'' agent directs agents as if they were limbs. There is no need to communicate as We accesses everyone's knowledge to coordinate agents perfectly. We can be formalized as a model of the underlying mind, composed of a set of mental states -- beliefs, desires, and intentions. Beliefs are the informational states of the mind, desires are the motivational states of the mind, and intentions are the deliberative states of the mind \cite{bratman1987intention}.

Theory of Mind (ToM) is a well-studied type of social reasoning \cite{wellman1992child}, with a powerful computational counterpart \cite{baker2009action}, providing a framework to process the mind for rational action planning and interpretation. Under this framework, agents aim to maximize their utility according to their mental states while minimizing costs of acting in the world \cite{gergely1995taking}. This process can also be reversed to understand others’ actions. In inverse planning, an observer uses Bayesian inference to infer the likely mental states that generates observed actions.

While We is a collective body of multiple cooperators, ToM traditionally models the mind of an individual. To accommodate, individual ToM has been extended to model the mind of We ($mind_{we}$) which contains \textit{joint} mental states: joint beliefs ($b_{we}$), joint desires ($d_{we}$), and joint intentions ($i_{we}$) \cite{kleiman2016coordinate, shum2019theory}. This allows reasoning about what we believe, what we want, and what we intend to do:

\begin{equation}\label{eq:MindProbability}
  P(mind_{we}) = p(b_{we})p(d_{we})p(i_{we}|b_{we}, d_{we})
\end{equation}

The problem with a centralized We is that it does not exist in reality; only You and I exist as individuals. However, We can be socially real, so long as each individual is imagining it. To imagine We, each agent tries to imagine how someone viewing the task from above would coordinate actors, taking on a ``bird’s eye perspective’’ \cite{tomasello2010origins} or ``view from nowhere’’ \cite{nagel1989view}. While the aim for each agent is to model the same We as the others, each agent may imagine a slightly different version of We. When these versions of We are synchronized, agents can coordinate smoothly.

Recent computational successes have shown that shared agency can help cooperators achieve complex goals without communication. In a gridworld paradigm, joint reasoning about high-level cooperation helps constrain joint actions in low-level coordination to collect joint rewards \cite{kleiman2016coordinate}. Additionally, in a cooking task, agents use joint reasoning to coordinate the execution of a high-level recipe step together \cite{wang2020too}. Shared agency also allows agents to bootstrap joint commitment to one of many goals, even under observation noise and model uncertainty \cite{tang2020bootstrapping}. In these examples, observed actions serve as feedback that can help align agents’ imagined We mind. In this paper, we call this modeling perspective the Imagined We (IW) to highlight that We is not real but instead, individually imagined.

As long as agents are able to synchronize their versions of the IW mind, shared agency is already a successful framework, even without communication. We take the perspective that communication is an even more powerful mechanism for coordinating perspectives. A key contribution of this work is to model communication as a means to synchronize perspectives by first building it up from an underlying shared agency framework. In addition, under this formalization, synchronization is aided by strong restrictions on what information is allowed in the IW: the IW considers only publicly shared information -- also called common ground \cite{clark1996using}. As agents take in new observations individually, communication can make this private knowledge mutually known. 

\subsection{Modeling Language Pragmatics}
Here we propose a mechanism for processing language. Language has been treated as rational and cooperative \cite{grice1975logic} where communicators aim to make themselves understood by speaking truthfully, relevantly, and efficiently. These ideas have been incorporated into a successful existing probabilistic model of language pragmatics: Rational Speech Acts (RSA) \cite{frank2012predicting, goodman2016pragmatic}. Social reasoning under RSA is recursive -- speakers and listeners can model each other, \textit{ad infinitum}. Deeper reasoning increases accuracy but at the cost of efficiency.

At the top level, a communicator aims to describe a referent or state of the world ($state$) with vocabulary ($signal$) that is true but possibly ambiguous. Signals are produced by a pragmatic speaker ($p_{sp}$) through a decision making process. A signal is chosen based on a noisy utility maximization (soft-max), where $\beta \in [0,\infty)$ represents the degree of rationality: 

\begin{equation}\label{eq:RSA_signalutilityGeneric}
P_{sp}(signal|state) \propto e^{\beta \mathbb{E}[U(signal, state)]}    
\end{equation}

RSA is a generalizable model of language production that provides the structure required to connect ambiguous language to concrete utilities. In RSA, a signaler’s utility is evaluated by modeling the pragmatic listener’s ($p_{lp}$) interpretation of that signal in terms of the states it could refer to. This requires a simpler, literal speaker ($p_{sl}$) model:

\begin{equation}\label{eq:RSA_signalutilityListener}
\begin{split}
\mathbb{E}[U(signal, state)]  & = p_{lp}(state|signal)  \\
 &\propto p_{sl}(signal|state)p(state)
\end{split}
\end{equation}

In this formulation, ``relevance'' is determined by the entire set of features in the environment, both ones the speaker highlights through language and the less direct ones that the speaker foregoes. It is not a trivial challenge to generalize utility, but here, it is ultimately limited to the ability to understand the referent of a signal as a target world state. In reality, language is often several degrees more indirect: communication can be used to explain not only to referent states (What?) but also social motivations (Why?) and interactions in the shared environment (How?).

A large part of non-referent context can be captured by modeling what an intelligent agent wants to do, why they want to do it, and how to accomplish it. While RSA language pragmatics and ToM decision-making in the physical environment have been formalized individually, they are rarely modeled in conjunction. The IW for Communication integrates these two perspectives in a novel way to broaden context, allowing a signal to be more indirect -- interpretable in terms of the IW’s joint beliefs, social motivations, \textit{and} actions.

\subsection{Deriving Signal Utilities from Action Utilities}
Here we explore how to formalize the utility of a signal under the IW. Utilities are measured with respect to a mind because they are subjective, socially dependent quantities. In the IW, this utility is in terms of a joint We mind. How can you derive a utility from the physical environment when the signal cannot actually change that environment? An action’s utility is measured by its consequence on the environment, so what are the direct consequences of a signal? Just as actions are designed to change the world, signals are designed to change the mind. If we can predict what consequence a signal has on our beliefs, desires, and intentions, we can connect this back to actions produced by rational planning in traditional ToM which have well defined utilities.

\begin{figure}[h!]
\centering
\includegraphics[width=.9\linewidth]{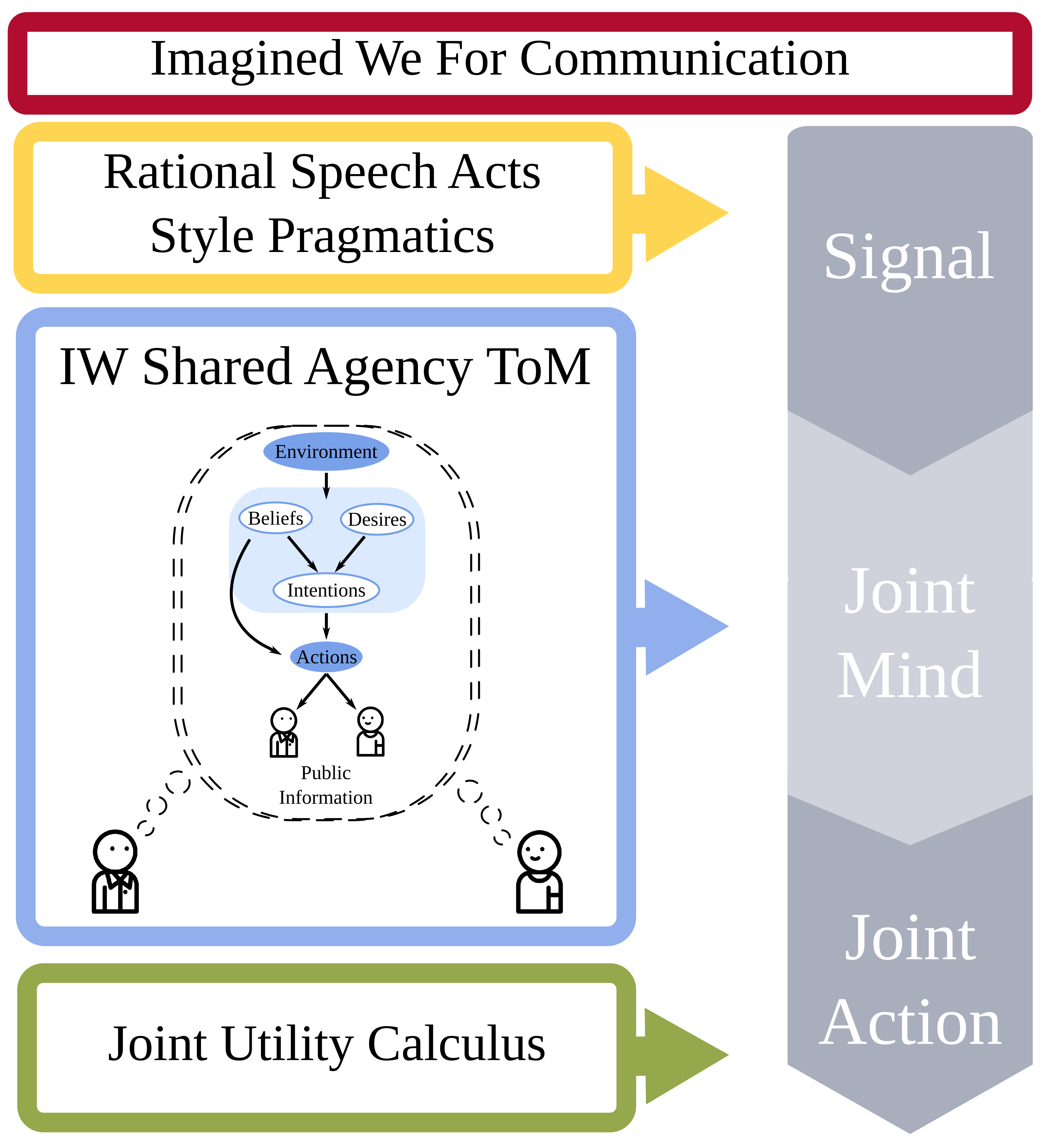}
\caption{Signals in the IW are processed through the mind and connected to joint action utilities.}
\label{fig:IW_frameworkFlow}
\end{figure}

We extend the IW for communication (see Fig.~\ref{fig:IW_frameworkFlow}) by leveraging RSA’s existing framework of cooperative, pragmatic language and redefining the utility of a signal to rely on shared agency ToM and interactions in the physical world. We connect these components by recognizing (1) signals change each IW, (2) minds produce predictable, rational joint actions under ToM reasoning, and (3) actions have well defined expected utilities which can be derived through joint planning.

Grounding a signal's value in a task where multiple agents interact in the environment is an approach that has been adopted in the AI communities. A signal's value is derived from its expected action consequences in a task in Recursive Mind Models \cite{gmytrasiewicz2001rational} which has been extended to sequential decision making problems with interactive partially observable Markov decision processes (I-POMDPs) \cite{gmytrasiewicz2005framework}. However, unlike the IW, I-POMDPs typically maximize an individual, self-interested reward. A cooperative alternative which assumes centralized training but individual observations and actions at execution is the decentralized POMDP (DEC-POMDP) \cite{oliehoek2016concise}. Critically, none of these models deal with the contextual flexibility of human-like communication. Instead, signal meanings must be pre-determined through a shared codebook that maps each signal onto its own thing. These approaches treat uncertainty in communication as noise in the signal channel instead of the intrinsic uncertainty of a signal that maps onto multiple referents.

\subsubsection{Joint Planning in Language:}
We have introduced ToM as a model of social cognition, but it also provides a rich framework for harnessing intuitive action costs and preference rewards \cite{jara2016naive, ullman2017mind}, which allows planning using the physical scene. Computational successes have emphasized the role of joint utility functions in morality \cite{kleiman2017learning} and team compositions \cite{shum2019theory}; furthermore, empirical evidence supports adult cooperators acting according to joint utilities \cite{torok2019rationality}. Joint utility considerations create a cooperative logic in communication where agents expect their partners to act fairly. This constrains a signal’s interpretation to things easier for the listener to accomplish. Even toddlers can disambiguate the referent of an ambiguous request for help through a basic joint utility consideration \cite{grosse2010}. In the scenario, two equivalent objects are equidistant from the toddler but near and far relative to the signaler. Children retrieve the far object more often when the signaler has free hands than in a second condition where their hands are occupied, demonstrating sensitivity to cooperative logic in requests. We use a similar joint utility calculus to ground signal utilities back to action utilities in the IW.

\section{Modeling Signaling under the Imagined We}
The IW framework allows us to resolve ambiguity in communication by imposing constraints from a rational utility maximization under cooperative logic. In this framework, the meaning of a signal is formally defined as the mind that generates the signal, the mind that the speaker would like the IW to have: a target. A signal should convey information that helps resolve uncertainty about what we believe, want, or intend to do. Uncertainty can be in any component of the mind and even generalize to a joint inference over multiple uncertain components. Here, we focus on uncertainty in goals for clarity. Thus the speaker rationally selects a potentially overloaded signal according to the true goal and target of inference, $goal_t$, not currently shared in the common ground:

\begin{equation}\label{eq:signalUtility}
P(signal|goal_t) \propto e^{\beta \mathbb{E}[U(signal, goal_t)]}    
\end{equation}

The utility of a signal is measured by looking at the utility of the outcome actions, weighted by how often those actions are expected to occur:

\begin{equation} \label{eq:signalUtilityThroughAction}
\mathbb{E}[U(signal, goal_t)] = \mathbb{E}_{P(a|signal)}[U(a, goal_t)]
\end{equation}

This framework actually serves to coordinate different perspectives: (1) A speaker predicts how a signal can change the IW mind (here, shared goal: $goal_{we}$) (2) and evaluates how good that change is according to their private observations of $goal_t$. The evaluation of $U(a, goal_t)$ includes the cost of taking $a$ and the reward if $a$ achieves $goal_t$.

Action prediction can be further broken down by connecting signals to actions via the mind. First, signals change the IW mind, making some goals more likely than others. Second, using the ToM likelihood function for action planning, we can calculate which actions are rational conditional on a given joint mind. We assume actions are conditionally independent from signals given the mind, captured by the intuition that signals can only influence actions through the mind:

\begin{equation} \label{eq:ActionProb|Signal}
P(a|signal)=\sum_{goal_{we}} P(goal_{we}|signal) P(a|goal_{we})
\end{equation}

Traditional ToM planning yields $P(a|goal_{we})$ and Bayesian inference allows us to measure how observing a signal will change the distribution of inferred goals. For the likelihood function we use a measure of consistency (Is this message truthful given the goal?), similar the literal speaker from RSA:

\begin{equation} \label{eq:BayesianMind}
P(goal_{we}|signal) \propto P(signal|goal_{we})P(goal_{we})
\end{equation}

The IW is a shared agency account of modeling language that is able to integrate different types of relevance -- language pragmatics and intuitive utilities -- to communicate rationally under different sources of ambiguity.

\section{Simulations}

\subsection{Task}
We test the IW in an gridworld task to demonstrate its ability to communicate successfully in a visual setting. This task combines feature overloading, which demands the language pragmatics studied by Frank \& Goodman (2012), but is enriched by a spatial scene that requires joint planning similar to the ambiguous helping from Grosse et al. (2010).

In this task, a signaler and a receiver cooperate to reach a target item in a nearly-fully observable gridworld environment. The key is the only bit of asymmetry of information between agents: only the signaler knows which item is the target. If either agent reaches the target, both receive a reward (+8); however, each step incurs a shared cost (-1). We calibrate the reward so that the expected utility of a signaler acting when it is better to ask for help is around zero. In play, the signaler acts first -- she may walk to and select an item (incurring the appropriate action cost), send a signal to her partner (free), or quit the trial (earning a utility of zero). Each item has two features: shape (circle, triangle, square) and color (red, green, purple). Signaling is costless but limited to conveying a single feature, adding ambiguity and increasing the chance it will refer to more than one item in the environment. If the first agent sends a signal (or walks to an incorrect item), the receiver then gets a turn. The trial ends after the receiver’s turn or the target item is reached, whichever comes first. Traveling to the wrong goal results in a negative utility equal to the number of steps taken to reach that item.

\begin{figure}[h!]
\centering
\includegraphics[width=.65\linewidth]{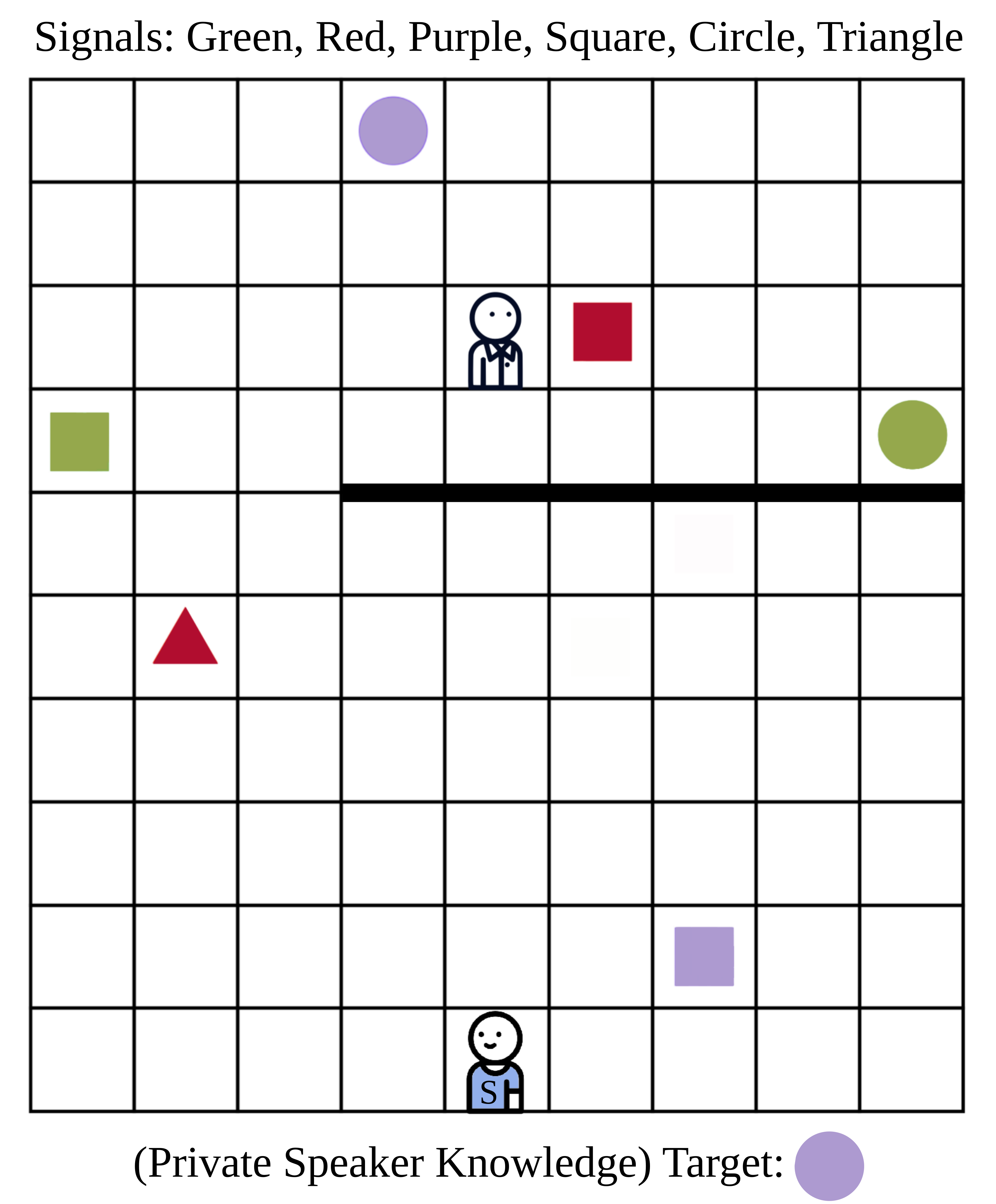}
\caption{Example trial setup with a barrier near the receiver.}
\label{fig:exTrial}
\end{figure}

In each trial, the set of shapes in the environment is randomly sampled and located; one is uniformly sampled to be the target, varying the optimal action and actor (see Figure \ref{fig:exTrial}). In addition, trials contain a physical barrier near the receiver which agents must go around to highlight the importance of joint planning. 

\subsection{Baseline Models}
We compare the IW model to two baselines and the central control optimal solution (CC). The optimal solution from the joint perspective is calculated with value iteration over the concatenation of the individual agents' action spaces. CC reflects how the two agents would rationally coordinate with perfect information: the ceiling of achievable utility.

The first baseline is a direct adaptation of RSA which we call acting RSA (aRSA). In its original formulation, RSA is intended to be a language only model; however, the current task involves reasoning about the cost of actions in the physical world. For this reason, the RSA signaler has the additional choice of whether they would like to perform an action, walking to an item instead of sending a signal. To make a fair comparison between these alternatives, the original signal utility (from Equation \ref{eq:RSA_signalutilityListener}) is augmented with the action utilities associated with particular inferences:

\begin{equation}
\mathbb{E}[U(signal, goal_t)] = \sum_{x \in items} p_{pl}(x|signal)U(a_{r} = x, goal_t)
\end{equation}

Assuming a rational receiver, the signaler evaluates the utility of the receiver traveling to an item ($a_{r} = x$) via value iteration and whether $x$ yields a reward (i.e. $ x = goal_t$). The pragmatic speaker takes the soft-max of the utilities of all signals, actions ending in an item, and quit option. The RSA receiver remains the same as before. This is different from the IW’s formulation of utility as it still lacks ``jointness’’ -- that is, action utility helps a signaler decide whether to communicate or act, but once a signal is sent, the receiver uses only pragmatic reasoning to decide what that signal means.

The second baseline model removes the pragmatic component of RSA to focus on joint utilities. Joint Utility (JU) agents use a joint utility calculation to apportion responsibility over items then uniformly send and interpret truthful signals in terms of the items they have deemed they are responsible for. The JU speaker takes the soft-max of the relative utilities of walking to the goal (do), partner walking to the goal (signal), and quitting. Similarly, a JU receiver, upon hearing a signal, constrains herself to consistent interpretations, weighing those interpretations according to their joint utilities.

Regardless of the model, a signaler will not communicate if it is more rational walk to the goal. To highlight each model’s use and understanding of communication, all simulation analysis is restricted to the trials where communication is optimal (i.e. the receiver goes to the target under the CC model and the achievable utility is greater than zero). 

\section{Simulation 1: Amount of Ambiguity}
We demonstrate the IW’s success in this cooperative communication task, even when signals are overloaded. The level of ambiguity is manipulated by increasing the number of potential target items in the environment (2-9 items). At six or more items, the target is guaranteed to be overloaded; that is, no matter what signal is chosen, it will be consistent with at least two items. We compare the utilities achieved by each model when faced with the exact same scenario.

\subsection{Results}
\begin{figure}[h!]
\centering
\includegraphics[width=.9\linewidth]{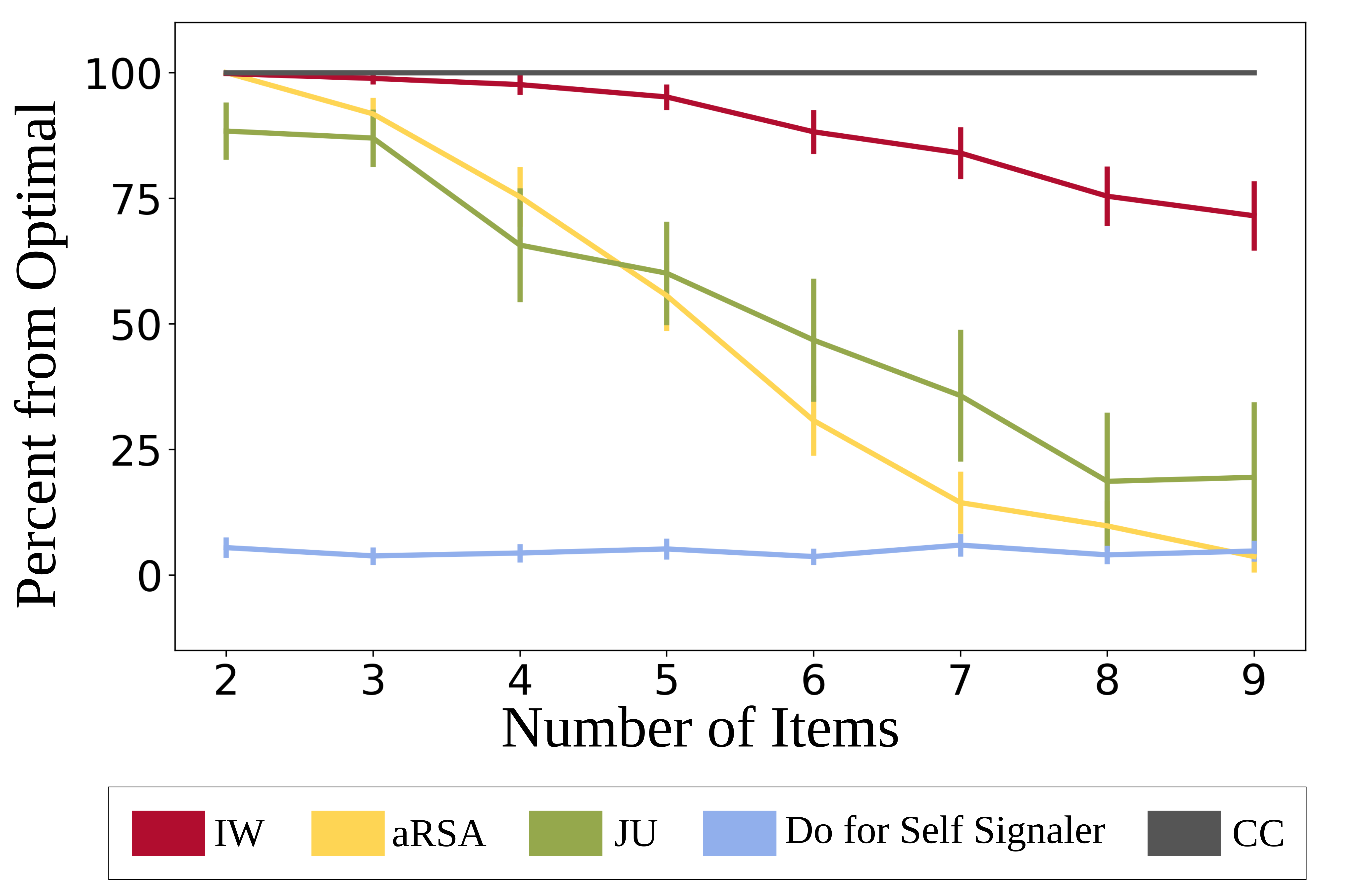}
\caption{Achieved utility (with 95\% CI) measured as the percent from optimal for each model under varying degrees of ambiguity. N=2000 trials per model. $\beta = 4$ for all models.}
\label{fig:Sim1_pctUtility}
\end{figure}

We look at the utilities achieved by each model, measured as a percent from the maximum possible as ambiguity increases (see Fig.~\ref{fig:Sim1_pctUtility}). When compared to always doing it for yourself (Do For Self Signaler), communication almost always leads to substantial gains in utility. This advantage disappears at high levels of ambiguity for the aRSA and JU models, but not for the IW. Across any number of items in the environment, the IW outperforms other baselines and only begins to deviate from the CC model when the uncertainty is very high. At the highest level of ambiguity (9 items) the IW achieves 71.5\% (CI: 64.6–78.4\%) of the optimal utility on average, while aRSA achieves 3.7\% (CI: 0.5–6.8\%) and JU achieves 19.4\% (CI: 4.4–34.4\%). This demonstrates how communication understanding is significantly enhanced by the integration of both cooperative pragmatics and joint utility reasoning in this task.

\begin{figure}[h!]
\centering
\includegraphics[width=.9\linewidth]{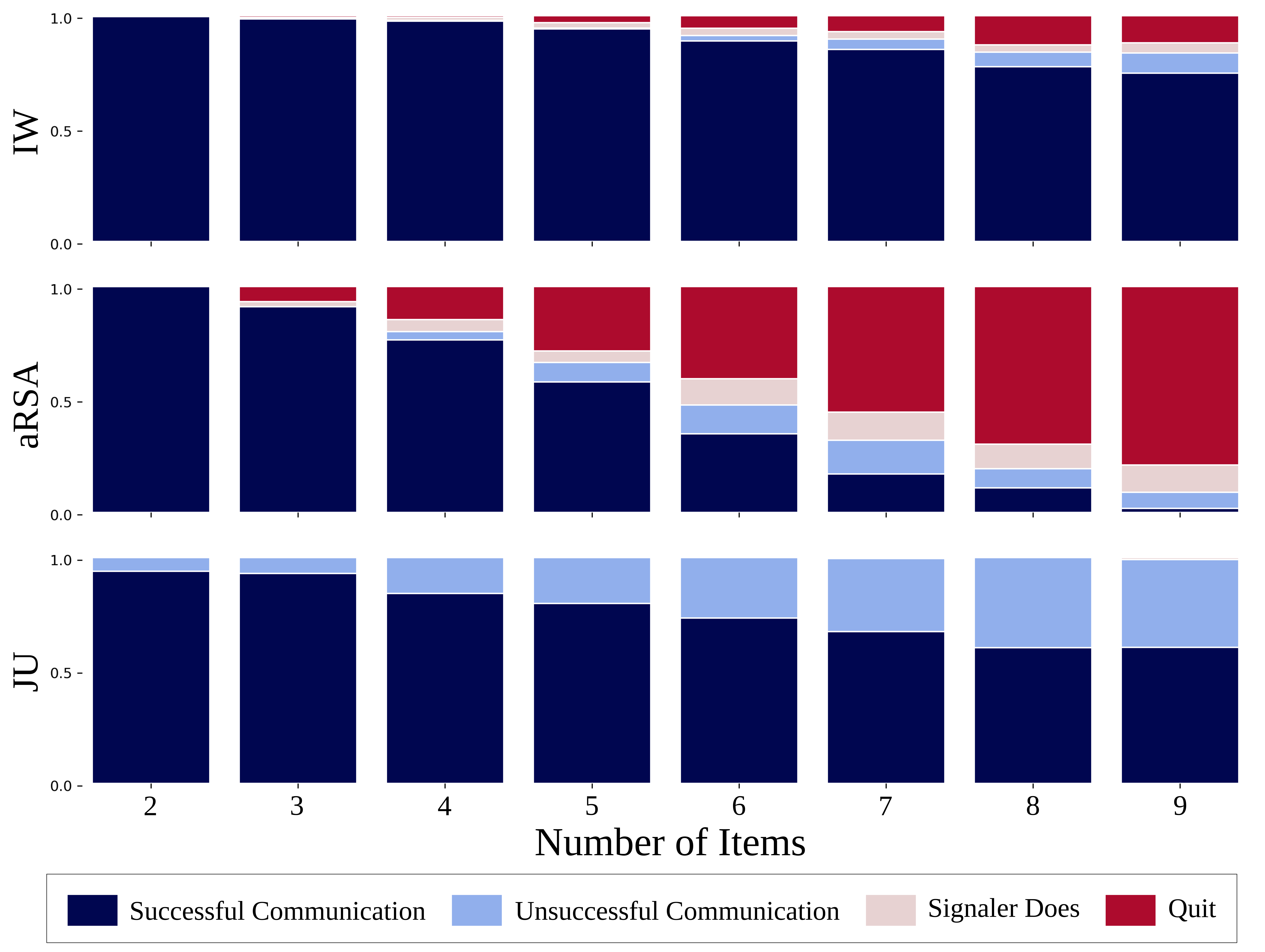}
\caption{Breakdown of agent behavior (as a proportion) for each model under varying levels of ambiguity. The behaviors are (1) Successful communication: the signaler communicates and the receiver goes to the correct goal (2) Unsuccessful communication: the signaler communicates and the receiver fails to choose the correct goal (3) Signaler does: the signaler forgoes communication and walks to the target (4) Quit: the signaler deems the trial too hard and skips the trial.}
\label{fig:Sim1_commBreakdown}
\end{figure}

We make a more fine-grained comparison between models to understand what contributes to differences in achieved utilities by breaking down model behaviors. As ambiguity increases, model behaviors diverge hugely (Fig.~\ref{fig:Sim1_commBreakdown}). The JU model always communicates (because analysis focuses on cases where communication is necessary to achieving the optimal utility). However, as ambiguity increases, this communication breaks down, and the receiver is increasingly less likely to correctly interpret the signal. In aRSA, the trend is dramatically different. Instead of unsuccessful communication, the signaler decides the uncertainty from the receiver’s interpretation is too large, leading to an unfavorable expected signaling utility. By nine items, the aRSA signaler is not able to successfully communicate at all.  The IW is able to combine these reasoning strategies to perform well across all levels of ambiguity; there is a much smaller decline in successful communication when the uncertainty is large. We see neither the breakdown of communication nor the huge increases in quitting behavior exhibited by the other models.

\section{Simulation 2: Level of Recursion}
One strength of the IW is that integrating the additional constraints from cooperative joint planning can often quickly resolve ambiguity, lessening reliance on deep recursion. This may provide a novel answer to why everyday pragmatic language is often quick and easy. Here we demonstrate this by looking at how performance changes as a function of deeper reasoning for the IW and aRSA at different levels.

RSA pragmatics involves recursion, which is increasingly expensive at each additional layer. A speaker starts with a literal model (level 0), which is used by a listener, which is then used by a pragmatic speaker (level 1). We can continue to build additional layers, each being more complex than the last. The IW can also handle this type of recursion; however, agents reason recursively about the joint IW mind, not each other. We compare the utilities achieved by different reasoning levels of speaker and receiver playing this task. 

In the IW, a joint utility calculation determines the portion of the environment where each agent is responsible for achieving the target. In addition to recursion levels, we test two different environmental configurations which change the joint utility dynamics: barrier near the receiver shown in Fig.~\ref{fig:exTrial} (RB) and barrier near the signaler -- the same barrier moved three grid spaces down (SB). By moving the barrier closer to the signaler, a larger portion of the environment becomes the receiver’s responsibility. This increases the difficulty by making the constraints from joint utility less likely to be useful for understanding a signal.

\subsection{Results}
\begin{figure}[h!]
\centering
\includegraphics[width=.9\linewidth]{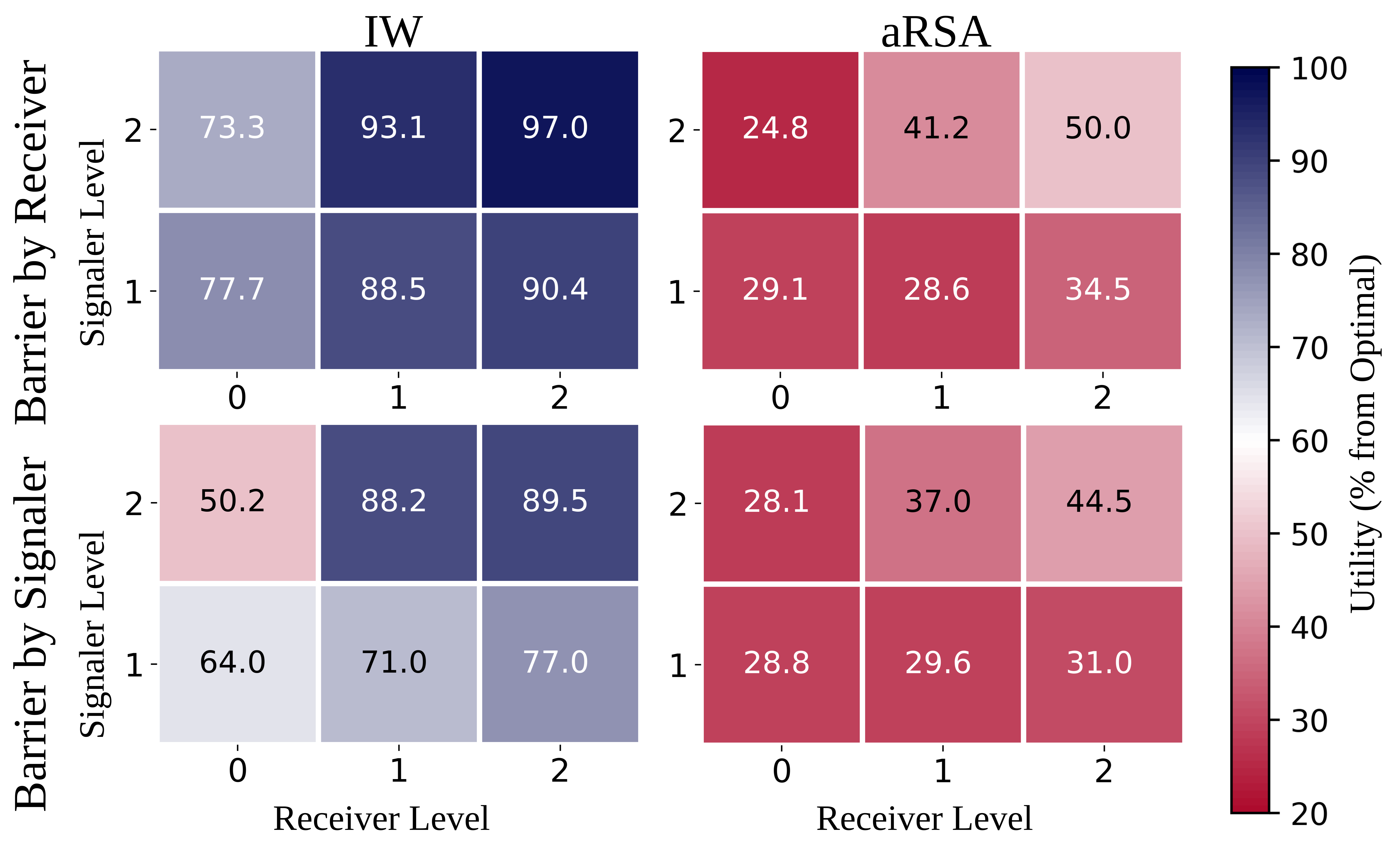}
\caption{Mean achieved utility (blue represents higher performance) for signaler and receiver pairs with different levels of reasoning. Shown are N = 500 cases where communication is optimal per modeling level pair. Number of items is fixed at 6, $\beta = 4$.}
\label{fig:Sim2_Recursion}
\end{figure}

In general, deeper recursive reasoning leads to an increase in performance. For both models, the most complex signaler/receiver pair (level-2 signaler, level-2 receiver) performs best regardless of the environment. When comparing the most complex pair to the simplest (level-1 signaler, level-0 receiver), the IW achieves an average of 19.3\% and 25.4\% boost in performance from recursion in the RB and SB conditions respectively; aRSA sees a 20.9\% and 15.6\% bump in performance. Within a signaler level, as the receiver does deeper reasoning the achieved utility tends to increase (see Fig.~\ref{fig:Sim2_Recursion}). This indicates that having an intelligent receiver is important to performing well on the task. Notably, for both models and environments, the worst performing pair is a level-2 signaler with a level-0 receiver. This could indicate that when the speaker expects their receiver to be reasoning more deeply than they actually are, this mismatch in expectations can be highly detrimental.

\begin{figure}[h!]
\centering
\includegraphics[width=.8\linewidth]{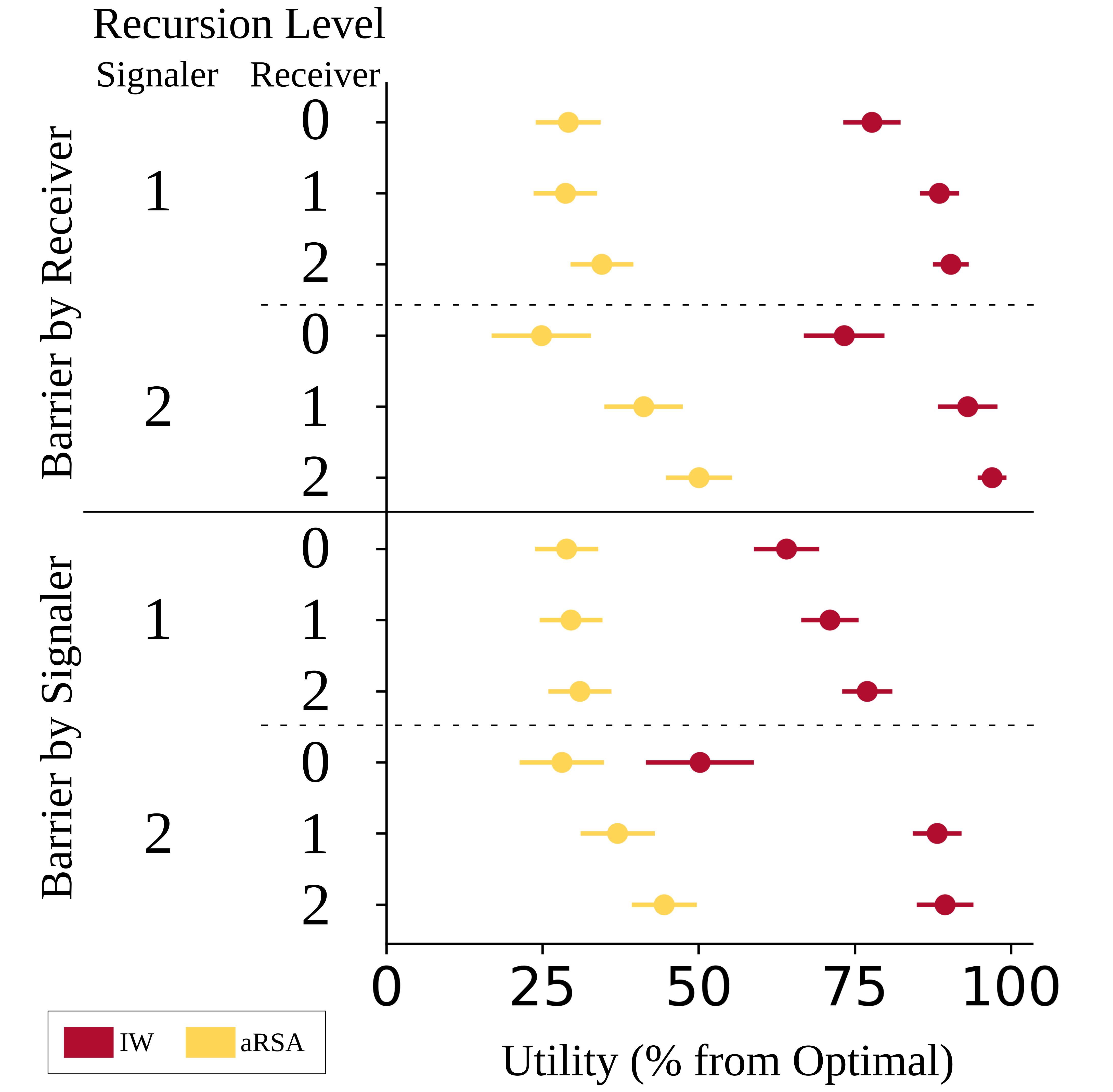}
\caption{Comparison of IW and aRSA mean achieved utilities with 95\% CIs.}
\label{fig:Sim2_Forest}
\end{figure}

At the same level of recursion, the IW always outperforms aRSA (see Fig.~\ref{fig:Sim2_Forest}), achieving up to twice the utility. In fact, the most complex reasoning under aRSA does worse than the simplest IW communicator pair. For IW the simplest reasoning achieves 77.7\% (CI: 73.1-82.3\%) and 64.0\% (CI: 58.8-69.3\%) of the optimal achievable utility in the RB and SB conditions respectively. In contrast, the most complex communicator pair under aRSA only reaches 50.0\% (CI: 44.7-55.3\%) and 44.5\% (CI: 39.3-49.7\%). This large performance difference indicates that the benefits of recursion are outweighed by the benefits of joint utility reasoning. Here much of the complex inferential burden of language can be pushed to a much simpler utility calculus. If these results align with future empirical behavioral data, it would provide evidence that everyday language does not need deep recursion to be sparse and successful.

Finally we can examine the effect of moving the barrier on performance. From a joint utility perspective, moving the barrier toward the receiver makes it harder to constrain the meaning of a signal using joint utility.  We find that performance for a communicator pair is better in the RB condition than in the SB condition in the IW (adjusted p $<$ .001 for all communicator pairs) but not in aRSA (adjusted p $>$ .05 for all communicator pairs), demonstrating the gains from joint utility reasoning.

\section{Conclusion}
The IW serves as a general framework of indirect and ambiguous signal production and understanding under multiple types of uncertainty. Our proposed modeling approach emphasizes a shared agency perspective that relies on existing computational infrastructure which has already successfully modeled cooperative coordination. Cooperative logic, pragmatic language reasoning, and affordable actions under a joint utility calculus constrain a signal's interpretation. Integrating these sources of context allow for fast, flexible signaling which helps remove the inferential burden from deep recursion. We demonstrate the strength of this modeling perspective by manipulating the amount of ambiguity in the environment as well as the depth of reasoning between interlocutors. By comparing performance of the IW to a set of baseline models, we demonstrate that the IW representation is more robust under uncertainty and does not require deep recursion to perform well -- allowing joint utility to do much of the heavy lifting in language understanding. These findings support an account of communication that is able to integrate and process multiple types of relevance for rich understanding despite sparse, indirect signaling.

\section{Acknowledgments}
This material is based upon work supported by the National Science Foundation
Graduate Research Fellowship Program under Grant No. DGE-1650604 to SS. In addition, this research was supported by DARPA PA 19-03-01 and ONR MURI project N00014-16-1-2007 to TG.

\bibliographystyle{apacite}

\setlength{\bibleftmargin}{.125in}
\setlength{\bibindent}{-\bibleftmargin}
\bibliography{2021IW_Cogsci}

\end{document}